\pdfoutput=1

\documentclass[11pt]{article}

\usepackage{authblk}
\usepackage[]{acl}

\usepackage{times}
\usepackage{latexsym}

\usepackage[T1]{fontenc}

\usepackage[utf8]{inputenc}

\usepackage{microtype}

\usepackage{makecell}

\usepackage{url}

\usepackage{float}
\usepackage{graphicx}
\usepackage[caption=false]{subfig}

\usepackage{multirow}
\usepackage{enumitem}

\usepackage{amssymb}

\newcommand{\sect}{\S}

\usepackage{color,soul}

\usepackage{todonotes}

\usepackage{footnote}
\usepackage{booktabs}

\newenvironment{fontppl}{\fontfamily{ppl}\selectfont}{\par} 
\definecolor{brightmaroon}{rgb}{0.76, 0.23, 0.28}

\newcommand*\samethanks[1][\value{footnote}]{\footnotemark[#1]}

\title{Proposition-Level Clustering
\\ for Multi-Document Summarization}

\author[1]{\bf Ori Ernst}
\author[1\Thanks{~~Equal contribution.}]{\bf Avi Caciularu}
\author[1\samethanks]{\bf Ori Shapira}
\author[2]{\bf Ramakanth Pasunuru}
\author[2]{\\ \bf Mohit Bansal}
\author[1]{\bf Jacob Goldberger}
\author[1]{\bf Ido Dagan}
{
\makeatletter
\renewcommand\AB@affilsepx{~~~~~~ \protect\Affilfont} \makeatother
\affil[1]{Bar-Ilan University}
\affil[2]{UNC Chapel Hill}
}
\affil[  ]{} 
\affil[  ]{\tt \{oriern, avi.c33, obspp18\}@gmail.com}
\affil[  ]{\tt \{ram, mbansal\}@cs.unc.edu}
\affil[  ]{\tt \{jacob.goldberger@, dagan@cs.\}biu.ac.il}

\date{}

\begin{document}
\maketitle

\begin{abstract}

Text clustering methods were traditionally incorporated into multi-document summarization (MDS) as a means for coping with considerable information repetition. Particularly, clusters were leveraged to indicate information saliency as well as to avoid redundancy. Such prior methods focused on clustering \textit{sentences}, even though closely related sentences usually contain also non-aligned parts. In this work, we revisit the clustering approach, grouping together sub-sentential \textit{propositions}, aiming at more precise information alignment. Specifically, our method detects salient propositions, clusters them into paraphrastic clusters, and generates a representative sentence for each cluster via text fusion.
Our summarization method improves over the previous state-of-the-art MDS method in the DUC 2004 and TAC 2011 datasets, both in automatic ROUGE scores and human preference.\footnote{
Our code and model are publicly available at \href{https://github.com/oriern/ProCluster}{https://github.com/oriern/ProCluster}.}

\end{abstract}

\section{Introduction}
\label{sec_intro}

Common information needs are most often satisfied by multiple texts rather than by a single one. Accordingly, there is a rising interest in Multi-Document Summarization (MDS) --- generating a summary for a set of topically-related documents. 
Inherently, MDS needs to address, either explicitly or implicitly, several subtasks embedded in this summarization setting. These include salience detection, redundancy removal, and text generation.
While all these subtasks are embedded in Single-Document Summarization (SDS) as well, the challenges are much greater in the multi-document setting, where information is heterogeneous and dispersed, while exhibiting substantial redundancy across linguistically divergent utterances. 
\begin{figure}[t]
    \centering
    \resizebox{\linewidth}{!}{
    \includegraphics{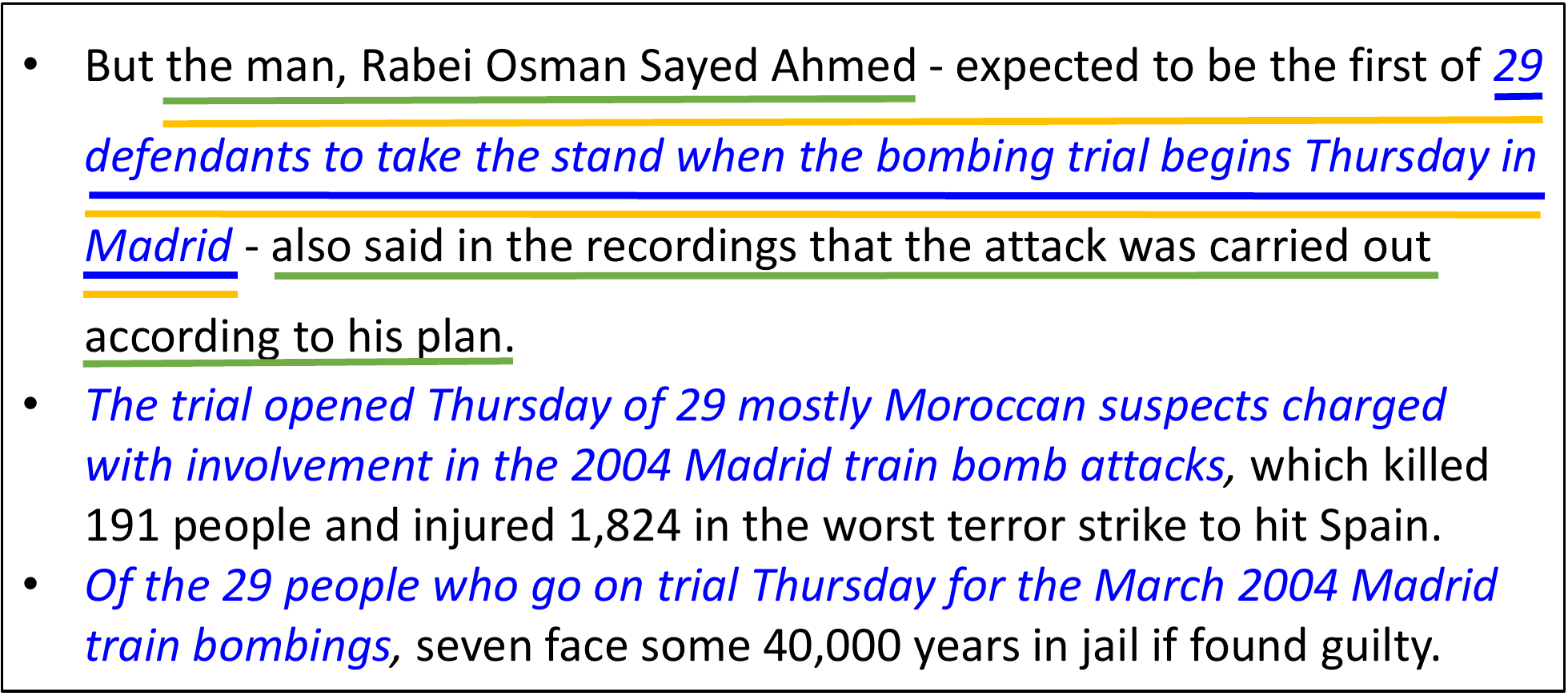}}
    \caption{An example of a cluster of \textcolor{blue}{\textit{propositions}}, shown within their source sentence \textcolor{black}{context}, from TAC 2011 (topic D1103). Clustering these as sentences would yield noisy unaligned information, however grouping together only the marked propositions keeps information alignment clean. The first sentence is illustratively divided into propositions, where only one of them is aligned to those in the other sentences.}
\label{fig:motivation_cluster}
\end{figure}

An appealing summarization approach that copes with these challenges, and is especially relevant for MDS, is clustering-based summarization. In such an approach, the goal is to cluster redundant paraphrastic pieces of information across the texts, which roughly convey the same meaning. Repetition of information across texts, as captured by paraphrastic clustering, typically indicates its importance, and can be leveraged for salience detection.
Moreover, representing a paraphrastic cluster may facilitate generating a corresponding summary that eliminates repetitions while fusing together complementary details within the cluster.

Traditionally, clustering-based approaches were widely used for summarization, mostly in extractive and unsupervised settings \citep{Radev2004CentroidbasedSO,  zhang-etal-2015-clustering, nayeem-etal-2018-abstractive}.
Notably, most of these works generated sentence-based clusters, which tend to be noisy since a sentence typically consists of several units of information that only partially overlap with other cluster sentences. As a result, such clusters often capture topically related sentences rather than paraphrases. Figure \ref{fig:motivation_cluster} exemplifies such a noisy cluster, which does contain paraphrastic propositions (marked in \textit{blue}) within their full sentences (marked in black).
Another line of research in summarization coped with such noisy sentence-based setting, and looked into the use of sub-sentential units for summarization, e.g., \citet{li-etal-2016-role} summarizes with Elementary Discourse Units (EDUs), while \citet{ernst-etal-2021-summary} endorse using Open Information Extraction
(OpenIE) -based propositions \citep{stanovsky-etal-2018-supervised} for summarization.

In this paper, we revisit and combine the clustering-based approaches along with sub-sentential setting, two research lines that were explored only individually and rather scarcely in recent years.
Specifically, we apply clustering-based summarization at the more fine-grained \textit{propositional} level, which avoids grouping non-aligned texts, yielding accurate paraphrastic clusters.
These clusters also provide better control over the generated summary sentences -- as the generation component is only required to fuse similar propositions.

Our model (\sect \ref{sec_dataCollection}) leverages gold reference summaries to derive training datasets for several summarization sub-tasks. First, salient document propositions were extracted, to train a salience model, by greedily maximizing alignment with the reference summaries. Then, an available
proposition similarity model, trained from summary-source alignments \citep{ernst-etal-2021-summary}, provides the  basis for agglomerative clustering  \citep{Ward1963HierarchicalGT}. Finally, we created training data for a BART-based model for sentence fusion  \cite{lewis-etal-2020-bart} by aligning reference summary propositions with source proposition clusters. Similar to many  other works, we leave inter-sentence coherence and sentence planning and ordering outside the scope of the current paper. Accordingly, our process produces a bullet-style summary of individual concise and coherent sentences.

Overall, our experiments (\sect \ref{sec_results}) show that this multi-step model outperforms strong recent end-to-end solutions, which do not include explicit modeling of propositions and information redundancy. 
To the best of our knowledge, our approach achieves state-of-the-art results in our setting on the DUC 2004 and TAC 2011 datasets, with an improvement of more than 1.5 and 4 ROUGE-1 F1 points respectively, over the previous best approach.
Finally, we also suggest (\sect \ref{sec_discussion}) that clustering-based methods provide ``explanations", or supporting evidence, for each generated sentence, in the form of the source cluster propositions (see an example in Table \ref{tab:cluster_examples}).

\section{Background and Related Work}
\label{sec_relatedWork}

\paragraph{Clustering-based summarization.}
Clustering-based summarization approaches typically involve salience detection while avoiding redundancy. One such approach clustered topically-related sentences, after which cluster properties were leveraged for rating sentence salience \citep{Radev2004CentroidbasedSO, Wang2008MultidocumentSV,  Wan2008MultidocumentSU}. Another approach rated sentence salience and clustered sentences simultaneously, iteratively improving the two objectives \citep{cai-etal-2010-simultaneous, Wang2011IntegratingDC, Cai2013RankingTC, zhang-etal-2015-clustering}. Recently, however, clustering methods have been gradually marginalized out, being replaced by neural techniques. More recently though, some approaches \citep{nayeem-etal-2018-abstractive, Fuad2019NeuralSF} presented abstractive clustering-based summarization, where topically-related sentences in each cluster are fused together to generate a summary sentence candidate. While most of previous clustering approaches operated at the noisy sentence level, in our work we present more accurate proposition-level clustering that eventually enhances summarization.

\begin{figure*}[t]
    \centering
    \resizebox{0.95\linewidth}{!}{
    \includegraphics{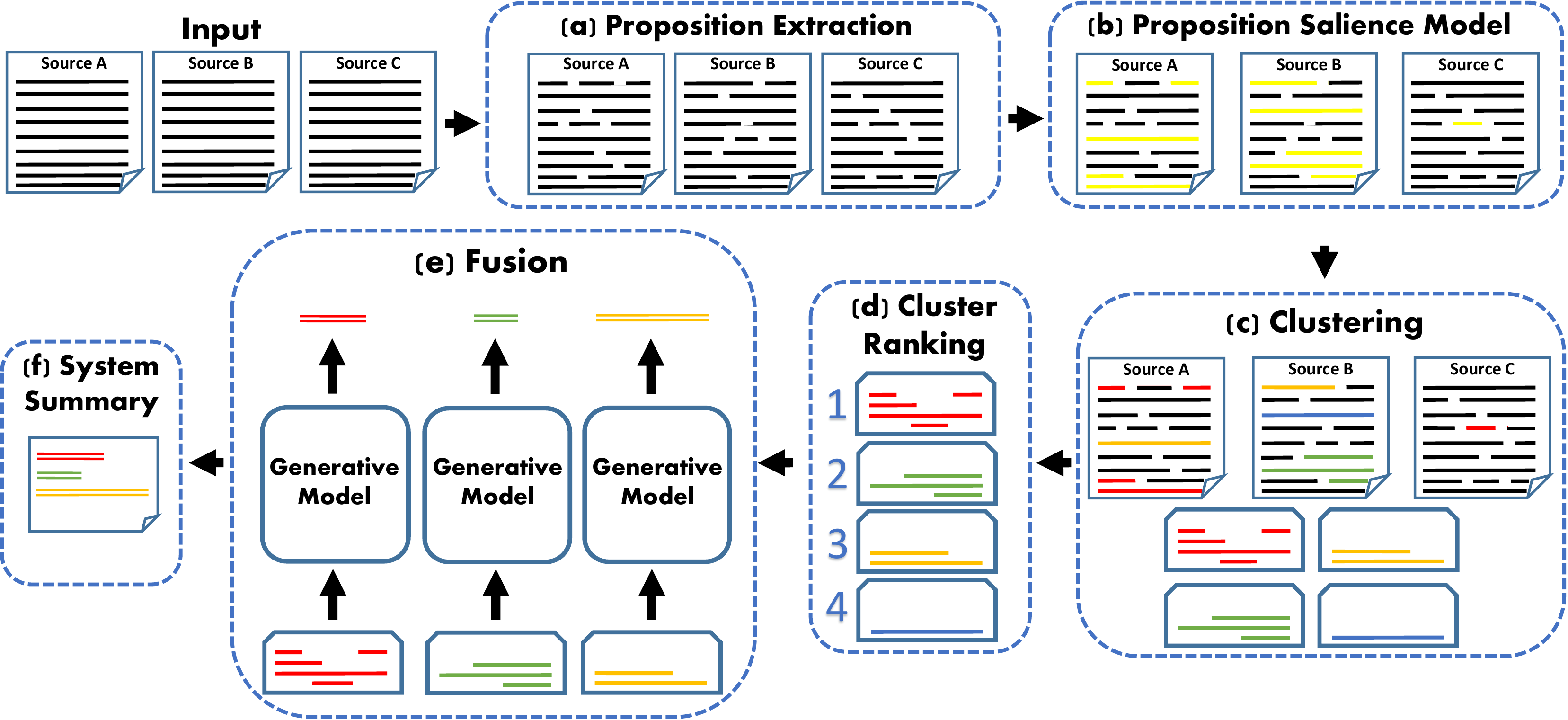}}
    \caption{Our multi-document summarization process. (a) All propositions are extracted \citep[OpenIE;][]{stanovsky-etal-2018-supervised} from the documents. (b) Propositions are classified by a salience score \citep[fine-tuned CDLM;][]{caciularu-etal-2021-cdlm-cross}. (c) Salient propositions are clustered \citep[fine-tuned SuperPAL;][]{ernst-etal-2021-summary}, forming groups of paraphrastic information units. (d) Clusters are ranked, as an indicator for information importance. (e) For each cluster, its propositions are fused \citep[fine-tuned BART;][]{lewis-etal-2020-bart} to generate a concise and coherent abstractive sentence. (f) The output summary is obtained as a bullet-style ranked list of the concise sentences.}
    \label{fig_method_process}
\end{figure*}

\paragraph{Sub-sentence units in summarization.}
While many summarization approaches extract full document sentences, either for extractive summarization or as an intermediate step for abstractive summarization, there are methods that operated the sub-sentential level. \citet{li-etal-2016-role, liu-chen-2019-exploiting}; and \citet{ xu-etal-2020-discourse} produced extractive summaries consisting of Elementary Discourse Units (EDUs) -- clauses comprising a discourse unit according to Rhetorical Structure Theory (RST) \citep{Mann1988RhetoricalST}. Such extractive approaches usually focus on content selection, possibly disregarding the inferior coherence arising from the concatenation of sub-sentence units. Accordingly, \citet{arumae-etal-2019-towards} established the highlighting task, where salient sub-sentence units are marked within their document to provide surrounding context.
Recently, \citet{cho-etal-2020-better} proposed identifying heuristically self-contained sub-sentence units for the highlighting task.

Abstractive approaches have been extracting sub-sentence units as a preliminary step for generation. Such units range from words \citep{lebanoff-etal-2020-cascade, gehrmann-etal-2018-bottom}, to noun or verb phrases \citep{bing-etal-2015-abstractive}, to OpenIE propositions \citep{pasunuru-etal-2021-efficiently}. In our work, we follow the same extract-then-generate pipeline, using OpenIEs \citep{stanovsky-etal-2018-supervised} as propositions. Since propositions are meant to contain single standalone facts consisting of a main predicate and its arguments, they are beneficial for grouping mostly overlapping paraphrases (unlike sentential paraphrases).
In addition, propositions extracted with OpenIE can be noncontiguous, while alternative options, like EDUs, are limited to contiguous sequences.

\section{Method}
\label{sec_dataCollection}

\newcolumntype{?}{!{\vrule width 2pt}}
\begin{table*}[t]
\setlength{\tabcolsep}{5pt}
\renewcommand{\arraystretch}{1.1}
\begin{scriptsize}
\begin{fontppl}

\begin{minipage}[b]{0.5\hsize}\centering
\begin{tabular}[t]{|p{2.99in}|}
\hline
\textbf{Cluster A}\\[1.1mm]
\textbullet\, The agreement will make Hun Sen prime minister and Ranariddh president of the National Assembly.
 \\[4.8mm]                        
\textbullet\, ...to a coalition deal...will make Hun Sen sole prime minister and Ranariddh president of the National Assembly.
 \\[4.8mm]

\textbullet\, The deal, which will make Hun Sen prime minister and Ranariddh president of the National Assembly...ended more than three months of political deadlock \\[6.8mm]

\textbullet\, Last week...Hun Sen's Cambodian People's Party and Ranariddh's FUNCINPEC party agreed to form a coalition that would leave Hun Sen as sole prime minister and make the prince president of the National Assembly.
 \\[9.8mm] 

\textbullet\, In a long-elusive compromise...opposition leader Prince Norodom Ranariddh will become president of the National Assembly
 \\[4.8mm] 
\hline
\hline
\textbf{Cluster B}\\[0.4mm]

\textbullet\, ...opposition party leaders Prince Norodom Ranariddh and Sam Rainsy are out of the country
 \\[3.2mm]

\textbullet\, Sam Rainsy and his then-ally Prince Norodom Ranariddh led an exodus of opposition lawmakers out of Cambodia
 \\[4mm]

\textbullet\, Opposition leaders Prince Norodom Ranariddh and Sam Rainsy...said they could not negotiate freely in Cambodia
 \\[4mm]

\textbullet\, Opposition leaders Prince Norodom Ranariddh and Sam Rainsy...citing Hun Sen's threats
 \\[3.85mm]

\hline
\end{tabular}
\end{minipage}
\hfill
\begin{minipage}[b]{0.5\hsize}\centering
\begin{tabular}[t]{|p{2.99in}|}
\hline
\textbf{Cluster C}\\[2mm]
\textbullet\, Hun Sen's Cambodian People's Party narrowly won the polls
\\[1.1mm]

\textbullet\, Hun Sen's ruling party narrowly won a majority in elections in July
\\[1.1mm]

\textbullet\, Hun Sen's Cambodian People's Party narrowly won the election.
\\[0.7mm]

\textbullet\, the ruling party narrowly won.
\\[1mm]
\hline
\hline
\textbf{Cluster D}\\[1mm]
\textbullet\, A series of negotiations to forge a new government
 \\[1.2mm]

\textbullet\, ...any...in deadlocked negotiations to form a government.
 \\[1.2mm]

\textbullet\, A series of negotiations to forge a new government have failed.
 \\[1.2mm]

\hline
\hline
\textbf{Cluster E}\\[1mm]
\textbullet\, \textcolor{red}{\textit{Hun Sen accused him}} of being behind a plot against his life.
 \\[1.2mm]

\textbullet\, Sam Rainsy...to take refuge in a U.N. office in September to avoid arrest after Hun Sen accused him of
 \\[4.2mm]

\textbullet\, Sam Rainsy...to avoid arrest after Hun Sen accused him of being behind a plot against his life.
 \\[3.2mm]

\hline
\hline
\textbf{Cluster F}\\[1mm]
\textbullet\, Hun Sen ousted Ranariddh in a coup.
 \\[1.2mm]

\textbullet\, The men served as co-prime ministers until Hun Sen overthrew Ranariddh in a coup last year.
 \\[3.5mm]

\textbullet\, Hun Sen overthrew Ranariddh in a coup last year.
 \\[1.2mm]

\hline

\end{tabular}

\end{minipage}

\begin{minipage}[b]{0.5\hsize}\centering
\begin{tabular}[t]{?p{2.95in}?}
\specialrule{2pt}{.05em}{.05em}
\textbf{ProCluster summary}\\[1.1mm]
\textbf{A.}\, The deal will make Hun Sen prime minister and Ranariddh president of the National Assembly\\[3.8mm]
\textbf{B.}\, The opposition party leaders Prince Norodom Ranariddh and Sam Rainsy are out of the country\\[3.8mm]
\textbf{C.}\, Hun Sen's Cambodian People's Party narrowly won the election.\\[1.4mm]
\textbf{D.}\, A series of negotiations to forge a new government failed.\\[1.4mm]
\textbf{E.}\, \textcolor{red}{\textit{The U.N. accused him}} of being behind a plot against his life.\\[1.4mm]
\textbf{F.}\, Hun Sen ousted Ranariddh in a coup last year.\\[1.4mm]
\textbf{G.}\, The opposition alleging widespread fraud and intimidation by the CPP\\[1.4mm]
\textbf{H.}\, The parties have refused to enter into a coalition with Hun Sen until their allegations of election fraud have been thoroughly investigated.\\[3.8mm]

\specialrule{2pt}{.05em}{.05em}

\hline
\end{tabular}
\end{minipage}
\hfill
\begin{minipage}[b]{0.5\hsize}\centering
\begin{tabular}[t]{?p{2.95in}?}
\specialrule{2pt}{.05em}{.05em}
\textbf{Reference Summary}\\[5mm]

Cambodia King Norodom Sihanouk praised formation of a coalition of the Countries top two political parties, leaving strongman Hun Sen as Prime Minister and opposition leader Prince Norodom Ranariddh president of the National Assembly.
\\

The announcement comes after months of bitter argument following the failure of any party to attain the required quota to form a government.
\\

Opposition leader Sam Rainey was seeking assurances that he and his party members would not be arrested if they return to Cambodia.
\\

Rainey had been accused by Hun Sen of being behind an assassination attempt against him during massive street demonstrations in September.
\\[13.5mm]
\specialrule{2pt}{.05em}{.05em}

\end{tabular}
\end{minipage}

\end{fontppl}
\end{scriptsize}
\caption{The proposition clusters and system and reference summaries for DUC 2004, topic D30001. Each summary sentence (lower left box) was fused from its corresponding cluster (top boxes) that also provides supporting source evidence. An example of an unfaithful abstraction is marked in \textcolor{red}{\textit{red}}.}
\label{tab:cluster_examples}
\vspace{-0.15in}
\end{table*}

This section first provides an overview of our method, followed by subsections describing its components. We follow previous clustering-based approaches, where text segments are first clustered into semantically similar groups, exploiting redundancy as a salience signal. Then, each group is fused to generate a merged sentence, while avoiding redundancy.
As we operate at the proposition-level, we first extract all propositions from the input documents (\sect \ref{subsec_method_OIE}). Then, to facilitate the clustering step, we filter out non-salient propositions using a salience model (\sect \ref{subsec_method_salience}). Next, salient propositions are clustered based on their semantic similarity (\sect \ref{subsec_method_clustering}).
The largest clusters, whose information was most repeated, are selected to be included in the summary (\sect \ref{subsec_method_rerank}).
Finally, each cluster is fused to form a sentence for a bullet-style abstractive summary (\sect \ref{subsec_method_generation}).
In addition, we provide an extractive version where a representative (source) proposition is selected from each cluster (\ref{subsec_method_ext_ver}). 
Overall, clustering explicit propositions induces a multi-step process that requires dedicated training data for certain steps. To that end, we derive new training datasets for the salience detection and the fusion models from the original gold summaries.
The full pipeline is illustrated in Figure \ref{fig_method_process}, where additional implementation details are in \sect \ref{appen_implementation_details} in the Appendix.

\subsection{Proposition Extraction}
\label{subsec_method_OIE}
Aiming to generate proposition-based summaries, we first extract all propositions from the source documents using Open Information Extraction (OpenIE) \citep{stanovsky-etal-2018-supervised}\footnote{\href{https://demo.allennlp.org/open-information-extraction}{https://demo.allennlp.org/open-information-extraction}}, following \citet{ernst-etal-2021-summary}. To convert an OpenIE tuple containing a predicate and its arguments into a proposition string, we simply concatenate them by their original order, as illustrated in Figure \ref{fig:OIE_example} in the Appendix.

\subsection{Proposition Salience Model}
\label{subsec_method_salience}
To facilitate the clustering stage, we first aim to filter non-salient propositions by a supervised model. To that end, we derive gold labels for proposition salience from the existing reference summaries.
Specifically, we select greedily propositions that maximize ROUGE-1\textsubscript{F-1} + ROUGE-2\textsubscript{F-1} against their reference summaries \citep{Nallapati2017SummaRuNNerAR, liu-lapata-2019-text} and marked them as salient.

Using this derived training data, we fine-tuned the Cross-Document Language Model (CDLM) \citep{caciularu-etal-2021-cdlm-cross} as a binary classifier for predicting whether a proposition is salient or not. Propositions with a salience score below a certain threshold were filtered out. The threshold was optimized with the full pipeline against the final ROUGE score on the validation set. All propositions contained in the clusters in Table \ref{tab:cluster_examples} are examples of predicted salient propositions. We chose to use CDLM as it was pretrained with sets of related documents, and was hence shown to operate well over several downstream tasks in the multi-document setting (e.g., cross-document coreference resolution and multi-document classification).

\subsection{Clustering}
\label{subsec_method_clustering}
Next, all salient propositions are clustered to semanticly similar groups. Clusters of paraphrastic propositions are advantageous for summarization as they can assist in avoiding redundant information in an output summary. Furthermore, paraphrastic clustering offers redundancy as an additional indicator for saliency, while the former salience model (\sect \ref{subsec_method_salience}) does not utilize repetitions explicitly.
To cluster propositions we utilize SuperPAL \citep{ernst-etal-2021-summary}, a binary classifier that measures paraphrastic similarity between two propositions. All pairs of salient propositions are scored with SuperPAL, over which standard agglomerative clustering \citep{Ward1963HierarchicalGT} is applied.
Examples of generated clusters are presented in Table \ref{tab:cluster_examples}.

\subsection{Cluster Ranking}
\label{subsec_method_rerank}
The resulting proposition clusters are next ranked according to cluster-based properties. We examined various features, listed in Table \ref{tab_cluster_ranking}, on our validation sets. 
The features examined include: average of ROUGE scores between all propositions in a cluster (\textit{`Avg. ROUGE'}), average of SuperPAL scores between all propositions in a cluster (\textit{`Avg. SuperPAL'}), average of the salience model scores of cluster propositions (\textit{`Avg. salience'}), minimal position (in a document) of cluster propositions (\textit{`Min. position'}), and cluster size (\textit{`Cluster size'}).

For each feature, (1) clusters were ranked according to the feature, (2) the proposition with the highest salience model score (\sect \ref{subsec_method_salience}) was selected from each cluster as a cluster representative, (3) the representatives from the highest ranked clusters were concatenated to obtain a system summary.
We also measured combinations of two features (\textit{`Cluster size + Min. position'} for example), where the first feature is used for primary ranking, and the second feature is used for secondary ranking in case of a tie. In all options, if a tie is still remained, further ranking between clusters is resolved according to the maximal proposition salience score of each cluster.
The resulting ROUGE scores of these summaries on validation sets are presented in Table \ref{tab_cluster_ranking}.\footnote{We also tried training a regression model on a mixture of features that should predict the ROUGE score of a proposition, but results were comparable. Bettering the ranking process is left for future work.}
We found that \textit{`Cluster size'} yields the best ROUGE scores as a single feature, and \textit{`Min. position'} further improves results as a secondary tie breaking ranking feature. Intuitively, a large cluster represents redundancy of information across documents thus likely to indicate higher importance.

\subsection{Cluster Fusion}
\label{subsec_method_generation}

Next, we would like to merge the paraphrastic propositions in each cluster, while consolidating complementary details, to generate a new coherent summary sentence. As mentioned, this approach helps avoiding redundancy, since  redundant information is concentrated separately in each cluster.

To train a cluster fusion model, we derived training data automatically from the reference summaries, by leveraging the SuperPAL model \citep{ernst-etal-2021-summary} (which was also employed in \sect \ref{subsec_method_clustering}). This time, the model is used for measuring the similarity between each of the cluster propositions (that were extracted from the documents) and each of the propositions extracted from the reference summaries. The reference summary proposition with the highest average similarity score to all cluster propositions was selected as the aligned summary proposition of the cluster. This summary proposition was used as the target output for training the generation model. Although these target OpenIE propositions may be ungrammatical or non-fluent, a human examination has shown that BART tends to produce full coherent sentences (mostly containing only a single proposition), even though it was finetuned over OpenIE extractions as target. Examples of coherent generated sentences can be seen in Table \ref{tab:cluster_examples}.

Accordingly, we fine-tuned a BART generation model \citep{lewis-etal-2020-bart} with this dedicated training data.
As input, the model receives cluster propositions, ordered by their predicted salience score (\sect \ref{subsec_method_salience}) and separated with special tokens.
The final bullet-style summary is produced by appending generated sentences from the ranked clusters until the desired word-limit is reached.

\begin{table}[!tbp]
\centering
    \resizebox{\linewidth}{!}{
    \begin{tabular}{c|c|c?c|c}
       \multirow{2}{*}{\textbf{Cluster Feature}} & \multicolumn{2}{c?}{DUC 2004} & \multicolumn{2}{c}{TAC 2011}\\ \cline{2-5}
       & R1 & R2 & R1 & R2 \\ \hline
    
    Avg. ROUGE  & 35.9 & 7.48  & 38.14 & 9.93 \\
    Avg. salience  & 35.5 & 7.98  & 41.18 & 12.55 \\
    Min. position  & 37.25 & 8.89 & 38.86 & 11.37 \\
    Avg. SuperPAL & 37.41 & 8.90  & 41.22 & 12.59 \\
     Cluster size & 37.58 & 9.01 & 41.35 & 12.49 \\
      Cluster size + Avg. SuperPAL & 37.54  & 8.96  & 41.45  & 12.71   \\
      Cluster size + Avg. salience & 37.77  & 9.09  & 41.44  & 12.62   \\
    Cluster size + Min. position & \textbf{38.05}  & \textbf{9.21}  & \textbf{41.68}  & \textbf{12.78}   \\

    \end{tabular}}
    \caption{ROUGE F1 results on validation sets when ranking clusters according to differing features (DUC 2004 is the validation set of TAC 2011 and vice versa). Two combined features means ranking on the first feature, and breaking ties with the second feature.}
    \label{tab_cluster_ranking}
    
\end{table}

\begin{savenotes}
\begin{table*}[!tbp]
\centering
    \resizebox{.9\textwidth}{!}{
    \begin{tabular}{cl|c|c|c?c|c|c}
    
       & \multirow{2}{*}{\textbf{Method}} & \multicolumn{3}{c?}{TAC 2011} & \multicolumn{3}{c}{DUC 2004}\\ \cline{3-8}
       
       & & R1 & R2 & RSU4 & R1 & R2 & RSU4 \\ \hline
       \multirow{7}{*}{\rotatebox[origin=c]{90}{absractive}}
    & Opinosis \citep{ganesan-etal-2010-opinosis} & 25.15 & 5.12  & 8.12 & 27.07 & 5.03  & 8.63\\
    & Extract+Rewrite \citep{song-etal-2018-structure} & 29.07 & 6.11 & 9.20 & 28.9 & 5.33 & 8.76\\
     & PG \citep{see-etal-2017-get} & 31.44 & 6.40 & 10.20 & 31.43 & 6.03 & 10.01\\
     & Hi-MAP\footnote{For the \textit{Hi-MAP} and \textit{MDS-Joint-SDS} approaches we present only DUC 2004 scores since TAC 2011 scores are not available for them.} \citep{fabbri-etal-2019-multi} & - & - & - & 35.78 & 8.90 & 11.43 \\
    & PG-MMR \citep{lebanoff-etal-2018-adapting} & 37.17  & 10.72  & 14.16   & 36.88  & 8.73  & 12.64\\
    & MDS-Joint-SDS\footnotemark[\value{footnote}]
    \citep{jin-wan-2020-abstractive} & - & - & - & 37.24  & 8.60  & 12.67   \\
     & ProCluster\textsubscript{abs} (Ours) & \textbf{41.45} & \textbf{12.75}  & \textbf{16.16}  & \textbf{38.71} & \textbf{9.62}  & \textbf{14.07}\\
     \hline
      \multirow{8}{*}{\rotatebox[origin=c]{90}{extractive}}
    & SumBasic \citep{VANDERWENDE20071606} & 31.58  & 6.06  & 10.06 & 29.48  & 4.25  & 8.64  \\ 
    & KLSumm \citep{haghighi-vanderwende-2009-exploring} & 31.23 & 7.07  & 10.56 & 31.04 & 6.03  & 10.23 \\
    & LexRank \citep{Erkan2004LexRankGL} & 33.10 & 7.50  & 11.13 & 34.44 & 7.11  & 11.19 \\

    & HL-XLNetSegs\footnote{The outputs of DPP-Caps \citep{cho-etal-2019-improving}, HL-XLNet and HL-Tree \citep{cho-etal-2020-better} were re-evaluated using author released output.} \citep{cho-etal-2020-better}  & 37.32 & 10.24  & 13.54 & 36.73 & 9.10  & 12.63 \\
    & HL-TreeSegs\footnotemark[\value{footnote}] \citep{cho-etal-2020-better} & 36.70 & 9.68  & 13.14 & 38.29 & \textbf{10.04}  & 13.57 \\
    & DPP-Caps-Comb\footnotemark[\value{footnote}] \citep{cho-etal-2019-improving} & 38.14 & 11.18  & 14.41  & 38.26 & 9.76  & 13.64 \\
    & RL-MMR \citep{mao-etal-2020-multi} & 39.65 & 11.44  & 15.02 & 38.56 & 10.02  & 13.80 \\
    & ProCluster\textsubscript{ext} (Ours)  & \textbf{40.98} & \textbf{12.40}  & \textbf{15.77} & \textbf{38.73} & 9.64  & \textbf{13.89} \\
    \hline
    & Oracle\textsubscript{prop}  & 49.65 & 21.82 & 23.19 & 46.49 & 16.16 & 18.76\\

    \end{tabular}}
    \caption{Automatic ROUGE F1 evaluation scores on the TAC 2011 \& DUC 2004 MDS test sets. Our solutions (ProCluster) improve over the previous state-of-the-art methods both in the abstractive and extractive settings. Notably, our \textit{abstractive} approach also surpasses the best \textit{extractive} ones.}
    \label{tab_main_res}
    
\end{table*}
\end{savenotes}

\subsection{Extractive Summarization Version}
\label{subsec_method_ext_ver}
To support extractive summarization settings, for example when hallucination is forbidden, we created a corresponding extractive version of our method. In this version, we extracted a representative proposition for each cluster, which was chosen according to the highest word overlap with the sentence that was fused from this cluster by our abstractive version.

\section{Evaluation}
\label{sec_results}

\subsection{Experimental Setup}
\paragraph{Datasets.}
We train and test our summarizer with the challenging DUC and TAC MDS benchmarks. Specifically, following standard convention \citep{mao-etal-2020-multi, cho-etal-2019-improving}, we test on DUC 2004 using DUC 2003 for training, and on TAC 2011 using TAC 2008/2009/2010 for training. These sets contain between 30 and 50 topics each.
For validation sets, we used DUC 2004 for the TAC benchmark and TAC 2011 for the DUC benchmark.

\paragraph{Automatic evaluation metric.}
Following common practice, we evaluate and compare our summarization system with ROUGE-1/2/SU4 F1 measures \citep{lin-2004-rouge}. Stopwords are not removed, and the output summary is limited to 100 words.\footnote{ROUGE parameters: -c 95 -2 4 -U -r 1000 -n 4 -w 1.2 -a -l 100 -m.} \footnote{Note that methods evaluated with ROUGE recall (instead of F1) or limited to 665 bytes (instead of 100 tokens) are not directly comparable to our approach.}

\subsection{Automatic Evaluation}
\label{subsec_results_automatic}

As seen in Table \ref{tab_main_res}, our abstractive model, denoted ProCluster\textsubscript{abs} for Propositional Clustering, surpasses all abstractive baselines by a large margin in all measures on both TAC 2011 and DUC 2004. 
Moreover, while the abstractive system scores were typically inferior to extractive system scores, ProCluster\textsubscript{abs} notably outperforms all extractive baselines in both benchmarks.
Overall, our ProCluster\textsubscript{abs} provides the new \textit{abstractive} MDS state-of-the-art score in this setting. In Figure \ref{tab:sys_summaries_examples} we present an example of a ProCluster\textsubscript{abs} system summary along with previous abstractive and extractive state-of-the-art system summaries and the reference summary.

As said in \sect \ref{subsec_method_ext_ver}, we also developed an extractive version, denoted ProCluster\textsubscript{ext}. As ProCluster\textsubscript{ext} selects document propositions that have the highest overlap with ProCluster\textsubscript{abs} sentences, ProCluster\textsubscript{ext} achieves similar scores to ProCluster\textsubscript{abs}, yielding the new \textit{extractive} MDS state-of-the-art results.

For comparison we selected strong baselines, including previous state-of-the-art in this setup, in both the extractive and abstractive settings. See in Appendix \sect \ref{subsec_baselines} for more concise details over each baseline.
For reference, we also present a proposition-based extractive upperbound for each dataset (\textit{Oracle\textsubscript{prop}}), where document propositions were selected greedily to maximize ROUGE-1\textsubscript{F-1} + ROUGE-2\textsubscript{F-1} with respect to the reference summaries.

\subsection{Ablation Analysis}

\begin{table*}[t]
\setlength{\tabcolsep}{5pt}
\renewcommand{\arraystretch}{1.1}
\begin{scriptsize}
\begin{fontppl}

\begin{minipage}[b]{0.5\hsize}\centering
\begin{tabular}[t]{|p{2.99in}|}
\hline
\textbf{RL-MMR}\\[1.1mm]

\textbullet\, An unknown number of cats and dogs suffered kidney failure and about 10 died after eating the affected pet food , menu foods said in announcing the north american recall .\\[8.8mm]
\textbullet\, Menu foods said saturday it was recalling dog food sold under 46 brands and cat food sold under 37 brands and distributed throughout the united states , canada and mexico .\\[8.8mm]
\textbullet\, Pet owners were worried that the pet food in their cupboards could be deadly after millions of containers of dog and cat food sold at major retailers across north america were recalled .\\[7.2mm]

\hline

\end{tabular}

\begin{tabular}[t]{?p{2.95in}?}
\specialrule{2pt}{.05em}{.05em}
\textbf{ProCluster summary}\\[1.1mm]

\textbullet\, The company announced the recall after receiving complaints that cats and dogs were suffering kidney failure.\\[1.8mm]
\textbullet\, Menu Foods recalled dog food sold under 48 brands and cat food\\[1.8mm]
\textbullet\, A major manufacturer of dog and cat food recalled 60 million containers of dog food.\\[1.8mm]
\textbullet\, The products were made by Menu Foods.The company\\[1.8mm]
\textbullet\, Cat food sold under 40 brands including Iams, Nutro and Eukanuba\\[1.8mm]
\textbullet\, The company began using wheat gluten from a new supplier\\[1.8mm]
\textbullet\, The 10 cats and dogs whose deaths have been linked to pet food\\[1.8mm]
\textbullet\, The food was distributed throughout the United States, Canada and Mexico.\\[1.8mm]
\textbullet\, Pet food sold under Wal-Mart, Safeway, Kroger and other store brands.\\[1.8mm]

\specialrule{2pt}{.05em}{.05em}

\end{tabular}
\end{minipage}
\begin{minipage}[b]{0.5\hsize}\centering
\begin{tabular}[t]{|p{2.99in}|}
\hline
\textbf{PG-MMR}\\[1mm]

\textbullet\, An unknown number of cats and dogs suffered kidney failure and about 10 died after eating the affected pet food , menu foods said in announcing the north american recall .\\[6.8mm]
\textbullet\, Menu foods , the ontario-based company that produced the pet food , said saturday it was recalling dog food sold under 40 brands including iams , nutro and eukanuba .\\[6.8mm]
\textbullet\, Menu foods is recalling only certain gravy-style pet food in cans and pouches it made from dec. 3 to march 6 .\\[3.8mm]
\textbullet\, Pet owners were worried that the pet food in their cupboard may be deadly after millions of containers of dog and cat food sold .\\[4.4mm]

\hline

\end{tabular}

\begin{tabular}[t]{?p{2.95in}?}
\specialrule{2pt}{.05em}{.05em}
\textbf{Reference Summary}\\[2mm]

\textbullet\, On Friday, March 16, 2007, Menu Foods of Streetsville, Ontario, began recalling 60 million containers of pet food after reports of ten animal deaths.\\[6.8mm]
\textbullet\, Menu's dog foods are sold under 48 brands and cat foods under 40 brands.\\[4.8mm]
\textbullet\, This company sells its products in the U.S., Canada and Mexico, and provides its products to 17 of the top 20 North American retailers.\\[5.8mm]
\textbullet\, The foods may have become contaminated by wheat gluten purchased from a new supplier which caused kidney failure in the animals.\\[6.8mm]
\textbullet\, The recalls bear code dates of 6339 through 7073.\\[2.8mm]
\textbullet\, The company will compensate owners of deceased animals.\\[2.6mm]

\specialrule{2pt}{.05em}{.05em}

\end{tabular}
\end{minipage}

\end{fontppl}
\end{scriptsize}
\caption{The system summaries and reference summary of topic D1104 in TAC 2011.}
\label{tab:sys_summaries_examples}
\vspace{-0.15in}
\end{table*}
\begin{table}[t]
\centering
    \resizebox{\linewidth}{!}{
    \begin{tabular}{c|l|c|c|c}
        & \textbf{method} & R1 & R2 & RSU4 \\ \hline
   \multirow{8}{*}{\rotatebox[origin=c]{90}{TAC 2011}}
    & Oracle\textsubscript{sent}  & 47.53 & 19.83  & 22.10 \\
     & Oracle\textsubscript{prop}  & 49.65 & 21.82 & 23.19 \\
     & Oracle\textsubscript{cluster-rep} & 43.40 & 14.61  & 17.46 \\
    & Oracle\textsubscript{ranking} & 46.38 & 17.59  & 19.88 \\
     \cline{2-5}
     & Salience\textsubscript{sent}  & 37.32 & 9.59  & 13.40 \\
     & Salience\textsubscript{prop}  & 39.92 & 11.53 & 15.12 \\
      & Salience\textsubscript{prop} + Clustering   & 41.05 & 12.40  & 15.73 \\
      & ProCluster\textsubscript{abs}  & 41.45 & 12.75  & 16.16  \\

    \specialrule{2.5pt}{1pt}{1pt}
    \multirow{8}{*}{\rotatebox[origin=c]{90}{DUC 2004}}  
    & Oracle\textsubscript{sent}  & 43.91 & 14.50  & 17.39 \\
     & Oracle\textsubscript{prop}  & 46.49 & 16.16 & 18.76 \\
         & Oracle\textsubscript{cluster-rep} & 39.74 & 10.76  & 14.56 \\
     & Oracle\textsubscript{ranking} & 43.70 & 12.92  & 16.43 \\
     \cline{2-5}
       & Salience\textsubscript{sent}  & 37.38 & 9.09  & 12.90 \\
     & Salience\textsubscript{prop}  & 37.73 & 8.97 & 13.18 \\
     & Salience\textsubscript{prop} + Clustering  & 38.41 & 9.09  & 13.56  \\
     & ProCluster\textsubscript{abs}  & 38.71 & 9.62  & 14.07  \\

    \end{tabular}}
    \caption{Ablation ROUGE F1 scores on TAC 2011 and DUC 2004. Each additional step in our multi-step method improves the output summaries. The Oracle results indicate the potential of our approach. Specifically, the benefit of summarizing on the proposition level is quite evident.}
    \label{tab_ablation}
    
\end{table}

To better apprehend the contribution of each of the steps in our pipeline, Table \ref{tab_ablation} presents results of the system when applying partial pipelines.

First, \textit{Salience\textsubscript{prop}} generates summaries simply consisting of the highest scoring document propositions, according to the CDLM-based salience model (\sect \ref{subsec_method_salience}).
We also trained the salience model on the sentence- rather than the proposition-level, and similarly generated summaries of salient sentences, denoted \textit{Salience\textsubscript{sent}}. The notable improvement of \textit{Salience\textsubscript{prop}} over \textit{Salience\textsubscript{sent}} in both datasets reveals the advantage of working at the proposition level for exposing salient information.
This observation is also apparent when comparing the proposition-based oracle (\textit{Oracle\textsubscript{prop}}) to the sentence-based oracle method (\textit{Oracle\textsubscript{sent}}).
The results indicate that proposition-based systems have a higher ROUGE upperbound across the board, supporting its merit for use in summarization.

Next, we would like to assess the contribution of the clustering step. Therefore, we applied Salience\textsubscript{prop} followed by clustering and ranking of clusters (Sections \ref{subsec_method_salience}, \ref{subsec_method_clustering} and \ref{subsec_method_rerank}), while leaving the fusion step aside. From each cluster we then select the proposition with the highest salience score to be in the system summary. In both datasets, the clustering stage provides added improvement, suggesting its contribution to our pipeline.

To further demonstrate the potential of our approach, we also present two additional oracle scores for extractive upperbound analysis. First, we examine the potential of optimally selecting cluster representatives for the summary. We greedily select a single representative per cluster following the original cluster ranking (\sect \ref{subsec_method_rerank}) that optimizes the overall ROUGE-1\textsubscript{F-1} + ROUGE-2\textsubscript{F-1} score of all selected representatives with respect to the reference summaries (\textit{Oracle\textsubscript{cluster-rep}}).
These results express the improvement comparing to our final model (\textit{ProCluster\textsubscript{abs}}), that a better cluster representative choice could produce, i.e., up to \textasciitilde 2 R-2 points in TAC 2011 and \textasciitilde 1 point in DUC 2004.

Another aspect to examine is the potential of enhanced cluster ranking. To that end, we first selected the highest salience-scoring proposition as a representative from each cluster. Then, we greedily selected representatives, one at a time, that maximized the overall ROUGE-1\textsubscript{F-1} + ROUGE-2\textsubscript{F-1} against the reference summaries. Effectively, this points to a greedily optimized cluster choice (\textit{Oracle\textsubscript{ranking}}). The potential improvement of better cluster ranking compared to our final model (\textit{ProCluster\textsubscript{abs}}) is hence up to \textasciitilde 5 R-2 points in TAC 2011 and \textasciitilde 3 points in DUC 2004. Indeed, our approach leaves cluster ranking improvement to future work.

Overall, we observe that all components of our multi-step approach are indeed effective for MDS, and that there is a great potential for further improvements within this architecture.

\subsection{Human Evaluation}
\label{subsec_results_human}
We further assessed our primary system, ProCluster\textsubscript{abs}, through manual comparison against PG-MMR and RL-MMR, which are state-of-the-art MDS systems in the abstractive and extractive settings (respectively). Crowdworkers on Amazon Mechanical Turk\footnote{\url{https://www.mturk.com}} were shown the summaries for a given topic from the three systems in arbitrary order, along with a corresponding reference summary. They were asked to rank the systems with respect to \textbf{Content} (content overlap with the reference), \textbf{Readability} (the degree to which a summary is readable and well-understood), \textbf{Grammaticality} (avoiding grammar errors), and \textbf{Non-Redundancy} (avoiding information repetition).
Focusing on evaluating our system, we extract from this ranking a pairwise comparison between our ProCluster\textsubscript{abs} and each of the two baseline systems, separately.
For each topic, this procedure was repeated for each of the four available reference summaries. Each such evaluation instance was judged independently by three workers, taking their majority vote for each pairwise comparison.

Table~\ref{tab_hum_res_rank} presents the results of these pairwise comparisons, showing the percentage of cases in which our system was preferred over each one of the two baselines, under each of the four evaluation criteria.
As can be seen, our system was favored in all cases, for both datasets. Furthermore, preference is almost always by a large margin, except for Non-Redundancy against RL-MMR, which avoids redundancy at a similar success level. Notably, as our clustering-based method is focused on improving content selection, the large gap in favor of ProCluster\textsubscript{abs} in the content criterion supports its advantage, consistently with our ROUGE-score advantage in the automatic evaluation (\sect \ref{subsec_results_automatic}).

While our summaries are (somewhat non-conventionally) structured as bullet-style lists of propositions rather than a coherent paragraph, evaluators preferred our style of summarization in terms of readability.
Moreover, as Table \ref{tab_abstractiveness} points out, ProCluster\textsubscript{abs} appears to be more abstractive than PG-MMR, as suggested by the reduced n-gram and sentence overlap with source documents. Specifically, about half of the system summary sentences of PG-MMR are fully copied, compared to about a quarter in our method. While the intensified abstractiveness of our summaries could have potentially hindered readability, our system was nevertheless preferred along this aspect as well.

Our approach leaves fertile ground for further improving readability by fusing several clusters together to generate sentences containing multiple propositions, and by developing sentence planning and ordering models. Compatible training datasets for these models can be derived out of the gold reference summaries, as was done in this work for the salience (\sect \ref{subsec_method_salience}) and fusion (\sect \ref{subsec_method_generation}) models.

\begin{table}[!tbp]
\centering
\resizebox{\linewidth}{!}{
    \begin{tabular}{cl|c|c|c|c}
      & \textbf{method} & Content & Read. & Grammar & Non-Red.\\ \hline
    \multirow{2}{*}{\rotatebox[origin=c]{90}{TAC}}
     & PG-MMR  & \textbf{93\%} & \textbf{84\%} & \textbf{81\%} & \textbf{72\%}\\
      & RL-MMR  & \textbf{82\%} & \textbf{70\%} & \textbf{74\%} & \textbf{52\%}\\

     \hline 
      \multirow{2}{*}{\rotatebox[origin=c]{90}{DUC}}
    & PG-MMR & \textbf{81\%} & \textbf{83\%} & \textbf{82\%} & \textbf{76\%} \\
    & RL-MMR & \textbf{70\%} & \textbf{72\%} & \textbf{71\%} & \textbf{54\%} \\

    \end{tabular}}
    \caption{Human pairwise comparisons between ProCluster\textsubscript{abs} and each of the two prior baseline systems, over the TAC 2011 and DUC 2004 datasets. The cells in a row show the percentage of cases in which our system was preferred over the corresponding  baseline, under each of the four evaluation criteria: content, readability, grammaticality and non-redundancy.}
    \label{tab_hum_res_rank}
    
\end{table}
\begin{table}[!tbp]
\centering
    \resizebox{\linewidth}{!}{
    \begin{tabular}{c|l|c|c|c|c}
       & \textbf{System} & \textbf{unigram} & \textbf{bigram} & \textbf{trigram} & \textbf{sent.} \\ \hline
   \multirow{3}{*}{\rotatebox[origin=c]{90}{TAC}}
    & PG-MMR & 98.36 & 94.42 & 91.97  & 50.11 \\
      & ProCluster\textsubscript{abs} & 99.08 & 91.40 & 81.07  & 24.39  \\
      & Ref. Summs. & 90.27  & 53.17 & 29.66  & 1.48 \\

    \specialrule{2.5pt}{1pt}{1pt}
    \multirow{3}{*}{\rotatebox[origin=c]{90}{DUC}}  
        & PG-MMR  & 98.34 & 94.99  & 90.91 & 50.82 \\
      & ProCluster\textsubscript{abs}  & 98.86 & 89.72  & 78.28 & 23.50 \\
      & Ref. Summs. &  88.41  & 44.27 & 18.65  & 0.13 \\

    \end{tabular}}
    \caption{Percentage of n-gram/sentence overlap between summaries and source documents in TAC 2011 and DUC 2004. Compared to PG-MMR, our system has substantially less sequential overlap, indicating its increased abstractiveness. Reference summaries are naturally highly abstractive.}
    \label{tab_abstractiveness}
    
\end{table}

\section{Paraphrastic Clusters as Summary Evidence}
\label{sec_discussion}

A unique advantage of a cluster-based summary is that each summary sentence is linked explicitly to a group of propositions from which the sentence was generated, in so providing an ``explanation'', or support evidence, for the output. These cluster explanations can expand the reader's knowledge and provide complementary facts from the nearby source context regarding the information from the generated sentence. Such a feature may be incorporated in interactive summarization systems, as applied in \citep{shapira2017intsum}, where a user can choose to expand on the facts within a sentence of the presented summary.

To assess the reliability of such feature, we verified that clusters indeed ``explain'' their generated sentences. To that end, we conducted a crowdsourced annotation, where a worker marked whether a cluster proposition mentions the main idea of its corresponding generated sentence. Each pair was examined by three workers, with the majority vote used for the final decision. On a random selection of 25\% of the clusters, we found that, on average, 89\% and 84\% of a cluster's propositions in DUC 2004 and TAC 2011 support their corresponding generated sentence, with an average cluster size of 3.4 and 4.8 propositions, respectively.

Furthermore, given this strong alignment of a cluster to its generated sentence,
a cluster facilitates effective verification of faithfulness of its corresponding generated abstractive sentence. Since the output sentence is based solely on its cluster propositions, the sentence's correctness can be verified against the ``explaining" cluster instead of against the full document set. An example of an unfaithful abstraction is marked in red in Table \ref{tab:cluster_examples}. To the best of our knowledge, this is the first attempt for efficient manual assessment of faithfulness in MDS.
We conducted a respective evaluation process, through crowdsourcing, to assess the faithfulness of our system summaries. A worker saw a cluster and its generated sentence and marked whether the sentence was faithful to its origin cluster or not.
Overall, this task cost a reasonable price of 240\$ for both the DUC 2004 and TAC 2011 datasets together.
Over the full test sets, the annotations showed that 80\% and 90\% of the DUC 2004 and TAC 2011 summary sentences, respectively, were faithful to their corresponding clusters.

\section{Conclusion}
\label{sec_conclusion}
We advocate the potential of proposition-level units as a cleaner and more accurate unit for summarization. To that end, we present a new proposition-level pipeline for summarization that includes an accurate paraphrastic propositional clustering component followed by fusion of cluster propositions, to generate concise and coherent summary sentences. Our proposed method outperforms state-of-the-art baselines in both automatic and human evaluation on the DUC and TAC MDS benchmarks. We provide an ablation study that indicates the benefit of each of the pipeline steps, as well as the potential for future improvement. Moreover, we demonstrate the utility of the clustering-based approach for providing source documents explanations and for manually validating summary faithfulness.

\section*{Acknowledgments}
\label{sec_acknowledgments}
The work described herein was supported in part by the PBC fellowship for outstanding PhD candidates in data science, Intel Labs, the Israel Science Foundation grant 2827/21, and by a grant from the Israel Ministry of Science and Technology.

\section*{Ethical Considerations}
\label{sec_ethical}
\paragraph{Computation.}
We ran on 3 GPUs for 20 minutes to finteune each of the salience model and the fusion model.

The summarization model runs 10 minutes on 4 GPUs to generate a summary. Most of the time is spent on the clustering step, in which we calculate the SuperPAL similarity score between all salient proposition pairs.

\paragraph{Dataset.}
The DUC 2003 and 2004 and TAC 2008-2011  datasets were acquired according to the required NIST guidelines ({\url{duc.nist.gov}}).

\paragraph{Crowdsourcing.}
All human annotations and evaluations conducted with crowdsourcing were compensated as a 12\$ per hour wage. We estimated the task payment by completing sample assignments and obtaining the average assignment time.

\bibliography{bibliography}
\bibliographystyle{acl_natbib}

\appendix

\section{Data Statistics}

\begin{table}[h]
    \centering

    \resizebox{\linewidth}{!}{
    \begin{tabular}{l|c|c|c}
         \textbf{Dataset} & \#Topics & \#Sents\textsubscript{per doc} &  \#Words\textsubscript{per doc}\\
        \hline
        DUC 2003  & 30 & 259   &6831\\
        DUC 2004  & 50 &  265  &6987\\
        TAC 2008-2010 & 138 &  237   & 5978\\
        TAC 2011 & 44 &  205  &5146\\
    \end{tabular}}

        \caption{Dataset statistics, including the number of document sets (i.e. topics) and the average number of sentences or words per document. Number of documents per topic is constant (10) for all datasets.}

        \label{table_data_stat}
\end{table}

\section{Implementation Details}
\label{appen_implementation_details}

\begin{figure*}[t]
    \centering
    \resizebox{\linewidth}{!}{
    \includegraphics{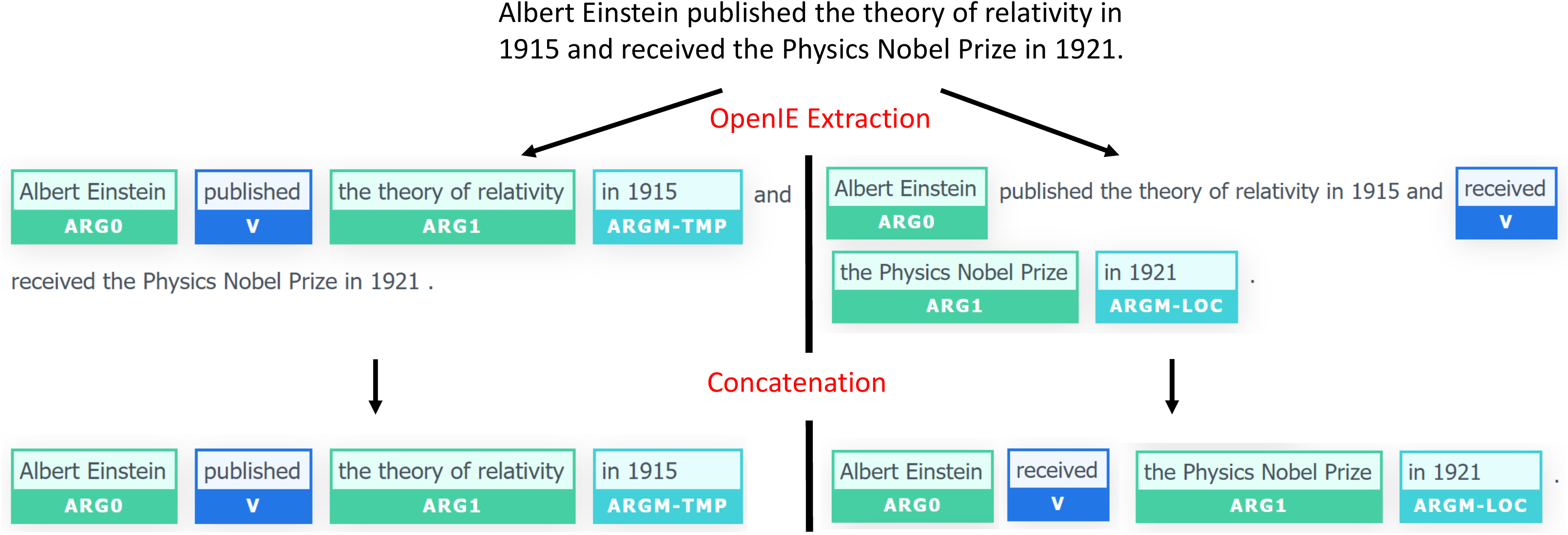}}
    \caption{An example of OpenIE spans extracted from a sentence. First, a sentence is divided to OpenIE tuples, including a predicate (verb) and its arguments. Then all predicates and their arguments are concatenated together to a full span. This illustration uses AllenNLP's Demo\footnote{\href{https://demo.allennlp.org/open-information-extraction}{https://demo.allennlp.org/open-information-extraction}}.}
\label{fig:OIE_example}
\end{figure*}

\subsection{Proposition Salience Model}
\paragraph{Datasets.}
For many previous summarization systems these benchmarks were insufficiently large enough for training their models. Consequently, they pretrained on a large scale summarization dataset, such as CNN/DailyMail \citep{Hermann2015TeachingMT}, and then finetuned on DUC/TAC datasets \citep[e.g.,][]{lebanoff-etal-2018-adapting, mao-etal-2020-multi}. In our case, we avoid external sources. However, as DUC training data is much smaller than TAC's (30 topics vs. 138), and it was apparently too small for the salience model training, we adopted the trained salience model for TAC benchmark (that was trained with TAC 2008-2010) as a pre-trained model and then finetuned it with DUC 2003.
Accordingly, validating the TAC benchmark using DUC 2004 during the salience model training causes data leakage since this model is later finetuned to test on the same DUC 2004. To avoid that, during the salience model training we used part of TAC 2010 that was omitted from training data, as a validation set (instead of DUC 2004).

\paragraph{Training Parameters.}
We trained the model for 10 epochs with learning rate of 1e-5 and batch size of 6 instances on 3 V100 GPUs (meaning effective batch size was 18).

\paragraph{Training.}

The CDLM model is fed with a proposition within its document and the other documents in the set. Specifically, since CDLM's input size is limited to 4,096 tokens, it is infeasible to feed the full document set as a long sequence. Therefore, following \citet{lebanoff-etal-2019-scoring}, only the first 20 sentences of each document are considered. Accordingly, a candidate proposition is input within its full document (up to 20 sentences), while other documents, ordered by their date, are truncated evenly and concatenated to fill the remaining space (9 sentences per document on average).

Each instance contains a proposition marked with start and end special tokens, within its multiple document context. A discontinuous proposition is marked with special tokens before and after each of its parts. In addition, sentence special token separators and document special token separators are used, as required for CDLM.

In order to reduce computation complexity, CDLM uses ``local attention" (of 512 tokens) for all tokens, while specific tokens are attended to all 4096 tokens (``global attention"). In our case, we assigned global attention to the CLS token and to the candidate proposition tokens, including their special tokens.

For classification, we have added a binary classifier layer on top of our CDLM. The classification layer gets the CDLM's CLS output representation concatenated to the sum of the CDLM output representations of the candidate proposition tokens:
\begin{equation}
    CLS \odot \sum_{i \in Prop} T_i
\end{equation}
where $T_i$ is the CDLM output representative of the $i$-th token, and $Prop$ contains the token indices of the candidate proposition.

As our proposition salience training dataset contains only a few positive (i.e., salient) propositions with respect to all propositions, it creates an unbalanced dataset that may strongly bias the model to give a negative prediction. To cope with this, we randomly filter out 60\% of the non-salient propositions, while over sampling salient propositions until the dataset becomes balanced.

\subsection{SuperPAL Usage}
In this work we used the SuperPAL model \citep{ernst-etal-2021-summary} as the similarity metric between propositions for the clustering step (\sect \ref{subsec_method_clustering}), and to create training data for the fusion model (\sect \ref{subsec_method_generation}). Originally, SuperPAL was tuned with a validation set that contains three topics from DUC 2004 (taken from the full validation set which also contains 7 additional topics, not from DUC 2004). In our setting, it may cause leakage since DUC 2004 is used as the test data. To avoid such leakage, we tuned SuperPAL again without using DUC 2004 topics at all (using the other 7 topics as a validation set).

\subsection{Cluster Ranking}
For computation time consideration, we set a maximum number of clusters to be selected for each topic. Since in most topics the 100-word limit is exceeded after 8-10 propositional sentences, we set the maximum number of clusters to 10. Accordingly, the 10 (or fewer) highest ranked clusters are selected for the summary of each topic.

\subsection{Fusion Model}
\paragraph{Training Parameters.}
We trained the model for 3 epochs with learning rate of 3e-5 and batch size of 10 instances on 3 V100 GPUs (meaning effective batch size was 30).

\section{Compared Methods}
\label{subsec_baselines}
We compare our method to several strong \textit{abstractive} baselines:
\textit{Opinosis} \citep{ganesan-etal-2010-opinosis} generates abstracts from salient paths in a word co-occurrence graph; \textit{Extract+Rewrite} \citep{song-etal-2018-structure} selects sentences using LexRank and generates for each sentence a title-like summary; \textit{PG} \citep{see-etal-2017-get} runs a Pointer-Generator model that includes a sequence-to-sequence network with a copy-mechanism; \textit{PG-MMR} \citep{lebanoff-etal-2018-adapting} selects representative sentences with MMR \citep{carbonell1998mmr} and fuses them with a PG-based model; \textit{Hi-MAP} \citep{fabbri-etal-2019-multi} is a hierarchical version of the PG model that allows calculating sentence-level MMR scores; \textit{MDS-Joint-SDS} \citep{jin-wan-2020-abstractive} is a hierarchical encoder-decoder architecture that is trained with SDS and MDS datasets while preserving document boundaries.

We additionally compare to several strong \textit{extractive} baselines:
\textit{SumBasic} \citep{VANDERWENDE20071606} extracts phrases with words that appear frequently in the documents; \textit{KLSumm} \citep{haghighi-vanderwende-2009-exploring} extracts sentences that optimize KL-divergence; \textit{LexRank} \citep{Erkan2004LexRankGL} is a graph-based approach where vertices represent sentences, the edges stand for word overlap between sentences, and sentence importance is computed by eigenvector centrality; \textit{DPP-Caps-Comb} \citep{cho-etal-2019-improving} balances between salient sentence extraction and redundancy avoidance by optimizing determinantal point processes (DPP); \textit{HL-XLNetSegs} and \textit{HL-TreeSegs} \citep{cho-etal-2020-better} are two versions of a DPP-based \textit{span} highlighting approach that heuristically extracts candidate spans by their probability to begin and end with an EOS token; \textit{RL-MMR} \citep{mao-etal-2020-multi} adapts a neural reinforcement learning single document summarization (SDS) approach \citep{chen-bansal-2018-fast} to the multi-document setup and integrates Maximal Margin Relevance (MMR) to avoid redundancy.

\section{Annotation Guidelines}

We used Amazon
Mechanical Turk\footnote{\url{https://www.mturk.com}} for all three crowdsource tasks with a list of 90 pre-selected workers from English speaking countries. These workers accomplished high quality work in other NLP-related tasks we have conducted in the past.

The crowdsourcing instructions of the tasks mentioned in \sect \ref{subsec_results_human} \& \ref{sec_discussion}  are as follows:

\subsection{General Summarization System Evaluation.}

Read the following four texts (Text A, B, C, and D) and answer the following questions.

\textbf{Text A:}\\
\textit{<Reference summary>}

\textbf{Text B:}\\
\textit{<System summary 1>}

\textbf{Text C:}\\
\textit{<System summary 2>}

\textbf{Text D:}\\
\textit{<System summary 3>}

\begin{itemize}

\item Which of the texts B, C, or D has \underline{the highest} content overlap with text A?

\item Which of the texts B, C, or D has \underline{the 2nd highest} content overlap with text A?

\item Which of the texts B, C, or D is \underline{the most} readable and well-understood?

\item Which of the texts B, C, or D is \underline{the 2nd most} readable and well-understood?

\item Which of the texts B, C, or D avoids grammar mistakes \underline{the best}?

\item Which of the texts B, C, or D avoids grammar mistakes \underline{the 2nd best}?

\item Which of the texts B, C, or D avoids information repetition \underline{the best}?

\item Which of the texts B, C, or D avoids information repetition \underline{the 2nd best}?

\end{itemize}

\subsection{Supporting Cluster Evaluation.}

Read the following two text spans, and answer the question below.

\textbf{Span Text A:}\\
\textit{<The generated sentence>}

\textbf{Span Text B:}\\
\textit{<A proposition from the cluster>}

\textbf{Is the main fact of Span Text A mentioned in Span Text B? (ignoring additional details)}\\
\indent Yes/No

\subsection{Faithfulness Evaluation.}

Read the following group of text spans A and text span B, and answer the questions below.
You can assume that all text spans in group A describe the same event, and therefore can be consolidated together to imply Text Span B.

\hspace{1mm}

\noindent\underline{Examples:}

\begin{enumerate}
\item
\textbf{Group of Text Spans A:}
\begin{itemize}
\item \textit{They arrested John.}
\item \textit{John was arrested.}
\end{itemize}

\textbf{Text Span B:}\\
\textit{The FBI arrested John}

\textbf{Is the Group of Text Spans A implies the fact in Text Span B?}

Text Span B add a detail that is not mentioned in A. Therefore the answer is No.

\item
\textbf{Group of Text Spans A:}
\begin{itemize}
\item \textit{there were 10-12 girls and 15 boys in the schoolhouse}
\item \textit{there were boys and girls in the schoolhouse}
\end{itemize}

\textbf{Text Span B:}\\
\textit{there were 1012 girls and 15 boys in the schoolhouse}

\textbf{Is the Group of Text Spans A implies the fact in Text Span B?}

Text Span B contradicts Group A (instead of 10-12 girls it says 1012 girls). Therefore the answer is No.
\end{enumerate}

\end{document}